\documentclass[runningheads]{llncs}

\usepackage{eccv}

\usepackage{eccvabbrv}
\usepackage{graphicx}
\usepackage{booktabs}
\usepackage{amsmath,amssymb}
\usepackage{comment} % Required for the commented out Ablation Study
\usepackage[accsupp]{axessibility}  % Improves PDF readability for those with disabilities.

\usepackage{hyperref}
\usepackage{orcidlink}

\begin{document}

% ---------------------------------------------------------------
\title{Debiasing Text-to-Image Evaluation via Implicit Cultural Alignment Reward Modeling} 

\titlerunning{Debiasing T2I Evaluation via Cultural Reward Modeling}

\author{Bo-An Chang\inst{1} \and Yu-Chih Chen\inst{2}}

\authorrunning{B.-A. Chang and Y.-C. Chen}
\institute{National Tsing Hua University, Hsinchu 300044, Taiwan \\
\email{boanzhang82@gapp.nthu.edu.tw}
\and National Yang Ming Chiao Tung University, Hsinchu 300093, Taiwan \\
\email{berriechen@nycu.edu.tw}}

\maketitle

\begin{abstract}
As Text-to-Image (T2I) systems rapidly advance, evaluating the cultural authenticity of synthesized content has become increasingly important for fair and trustworthy generative AI. Existing T2I evaluation metrics and multimodal judges often rely on visual-semantic representations that underrepresent implicit cultural norms, leading to biased preference judgments and the omission of fine-grained cultural cues. In addition, visual question answering (VQA)-based evaluators typically depend on autoregressive text generation, which limits their scalability for real-time reward modeling. To address these limitations, we introduce an Implicit Cultural Alignment Reward Model built upon a lightweight 4.2-billion-parameter Multimodal Large Language Model (MLLM). Our framework integrates an Implicit Cultural Probe with a Skip-connection Cross-Attention (SkipCA) mechanism, enabling late-stage semantic features to directly attend to early-stage visual representations and better preserve culturally salient details. Evaluations on 3,323 challenging and carefully curated image pairs from the CulturalFrames benchmark show that our approach achieves 83.49\% pairwise accuracy, with Pearson and Kendall correlation coefficients of 0.5268 and 0.3749, respectively, outperforming representative vision-language metrics and MLLM-based evaluators. Moreover, by bypassing autoregressive text generation, our model processes each evaluation in 0.21 seconds under our local inference setup, achieving a $10\times$ speedup over standard VQA-based evaluators. These results suggest that the proposed reward model can provide an efficient and culturally aware scalar signal for preference optimization pipelines such as Reinforcement Learning from Human Feedback and Direct Preference Optimization. Additional resources are available on our project page at \url{https://bensonch1214.github.io/Implicit_Cultural_Alignment/}.
\vspace{-3mm}
\end{abstract}

\section{Introduction}

As text-to-image (T2I) generation technologies become increasingly prevalent, evaluating the cultural authenticity of synthesized content has become a critical dimension of fair and trustworthy generative AI. Although recent T2I systems can generate visually compelling images from natural-language prompts, their outputs often reflect biases inherited from large-scale web data, where Western-centric visual norms and demographic stereotypes are disproportionately represented~\cite{jha2024visage,elsharif2025cultural}. As a result, cross-cultural prompts may be interpreted, synthesized, and evaluated according to dominant visual conventions, leading to culturally misaligned outputs and the omission of context-specific symbolic details.

Following \textit{CulturalFrames}, user expectations in cross-cultural image generation can be broadly categorized into ``explicit expectations'' and ``implicit expectations''~\cite{nayak2025culturalframes}. Explicit expectations refer to visual elements directly specified in the prompt, such as generating ``a Chinese family'' or ``Peking duck.'' In contrast, implicit expectations involve unspoken cultural norms, practices, objects, and contextual details that are commonly understood within a specific cultural setting. For example, a Chinese dining scene may be expected to feature chopsticks rather than forks in many traditional dining contexts, while a Hindu wedding ceremony typically involves a sacred fire. These details are often not explicitly stated in the prompt, yet their absence or incorrect rendering can substantially affect the perceived cultural authenticity of the generated image.

Existing automated evaluation metrics are ill-equipped to diagnose such implicit cultural misalignments. Traditional vision-language metrics, such as CLIPScore~\cite{hessel2021clipscore}, primarily rely on global image-text similarity and therefore have limited sensitivity to fine-grained, long-tail cultural artifacts. Visual question answering (VQA)-based and MLLM-based metrics, such as VQAScore~\cite{lin2024vqascore}, VIEScore~\cite{ku2024viescore}, T2IEval~\cite{huang2023t2icompbench}, and EvalAlign~\cite{hu2024evalalign}, provide stronger reasoning capabilities but remain costly at inference time due to autoregressive text generation. Moreover, in many decoder-only MLLM-based evaluators, fine-grained visual evidence can be diluted as visual and textual representations are compressed into deeper semantic states, making it difficult to preserve subtle cultural cues during reward prediction.

To address these challenges, we propose an Implicit Cultural Alignment Reward Model for T2I evaluation. Built upon a lightweight 4.2B-parameter MLLM backbone, our framework introduces an Implicit Cultural Probe to guide cultural feature inference and a Skip-connection Cross-Attention (SkipCA) mechanism to reconnect late-stage semantic representations with early-stage visual tokens. This design enables the reward model to revisit low-level visual evidence before producing a scalar cultural alignment score. The model is trained with a Bradley-Terry pairwise ranking loss~\cite{bradley1952rank}, following recent preference-based reward modeling practices for T2I generation.

The main contributions of this paper are summarized as follows:
\begin{itemize}
\item We propose an Implicit Cultural Alignment Reward Model for T2I evaluation, integrating an Implicit Cultural Probe and a SkipCA module to better preserve fine-grained cultural cues during multimodal reward prediction.
\item We show that the proposed model improves cultural alignment evaluation on 3,323 challenging image pairs from the CulturalFrames benchmark, achieving 83.49\% pairwise accuracy and outperforming representative evaluators, including GPT-4o at 72.54\% and VQAScore at 76.83\%.
\item We demonstrate that the proposed non-autoregressive reward prediction framework substantially improves inference efficiency, reducing evaluation time to 0.21 seconds per image (measured on a single NVIDIA RTX 5090 GPU) and providing a scalable scalar signal for downstream preference optimization methods such as Reinforcement Learning from Human Feedback (RLHF)~\cite{ouyang2022training,christiano2017deep} and Direct Preference Optimization (DPO)~\cite{rafailov2023direct}.
\end{itemize}

\section{Related Work}

\subsection{General Automatic Text-to-Image Evaluation}
Early assessment of T2I generation often relied on perceptual and distributional metrics such as Inception Score (IS)~\cite{salimans2016improved} and Fréchet Inception Distance (FID)~\cite{heusel2017gans}, which measure image quality and distributional fidelity. However, these metrics do not directly assess whether a generated image is semantically aligned with the input prompt. Consequently, recent research has shifted toward automated evaluators capable of assessing text-image alignment, visual fidelity, and human preference.

\noindent\textbf{CLIP- and preference-based metrics.}
Prior methods such as CLIPScore~\cite{hessel2021clipscore}, PickScore~\cite{kirstain2023pick}, HPSv2~\cite{wu2023human}, and ImageReward~\cite{xu2023imagereward} either employ contrastive vision-language representations or incorporate human preference data to estimate image-text alignment and perceptual quality. These metrics are efficient and scalable, making them widely used for evaluating T2I models. However, their reliance on global image-text embeddings can limit their sensitivity to fine-grained visual details, especially long-tail cultural artifacts that may not be explicitly described in the prompt. Furthermore, the preference datasets used to train or calibrate these evaluators may reflect demographic and cultural biases in the underlying data collection and annotation processes, potentially affecting their reliability in cross-cultural evaluation settings.

\noindent\textbf{MLLM-based evaluators.}
To improve the granularity of T2I evaluation, recent approaches leverage MLLMs. These methods can be broadly grouped into two paradigms. The first category comprises VQA-based evaluators, such as VQAScore~\cite{lin2024vqascore}, VIEScore~\cite{ku2024viescore}, T2IEval~\cite{huang2023t2icompbench}, and EvalAlign~\cite{hu2024evalalign}, which use prompt engineering to guide MLLMs in generating rationales, categorical judgments, or structured scores. While these methods provide richer reasoning capabilities, their reliance on autoregressive generation introduces substantial inference cost, making them less suitable for large-scale evaluation or dynamic training loops such as RLHF. The second category comprises token-probability-based methods~\cite{xu2023imagereward,hu2023tifa}, which formulate evaluation as the likelihood of predicting specific decision tokens, such as ``Yes'' or ``Good''. Although more efficient than full rationale generation, these methods remain tied to discrete verbal labels and may have limited sensitivity to subtle preference differences between visually similar samples. In contrast, our approach directly predicts a continuous scalar reward and optimizes it with a Bradley-Terry pairwise ranking loss, enabling efficient and fine-grained preference-based evaluation.

\subsection{Culture-Specific and Factual Evaluation}
As generative models are increasingly deployed across global contexts, cultural authenticity and factual grounding have become important dimensions of fair and trustworthy AI evaluation. However, general-purpose metrics are often not designed to assess culturally specific visual norms, implicit contextual expectations, or geographically grounded artifacts.

\noindent\textbf{Cultural alignment and implicit biases.}
Recent studies have examined cultural representation, stereotypes, and bias in text-to-image generative models. Frameworks such as CUBE~\cite{kannen2025cube} decompose cultural competence into dimensions such as awareness and diversity, highlighting the difficulty of generating culturally faithful content for underrepresented regions and communities. Building on this direction, CulturalFrames~\cite{nayak2025culturalframes} distinguishes between ``explicit'' and ``implicit'' cultural expectations. Explicit expectations correspond to visual elements directly specified in the prompt, whereas implicit expectations involve culturally salient details that may be unstated but still important for perceived authenticity. For example, a Hindu wedding scene may be expected to include a sacred fire, while a traditional Chinese dining scene may be expected to contain culturally appropriate utensils. Such cases show that a generated image may satisfy the surface-level prompt while still missing culturally meaningful visual evidence.

\noindent\textbf{Knowledge-driven factual metrics.}
To address the limited cultural and factual knowledge encoded in general-purpose evaluators, recent work has explored knowledge-driven and culture-specific evaluation frameworks. For example, CAIRe~\cite{yayavaram2025caire} uses visual entity linking to connect generated image content with external knowledge bases, supporting more grounded assessment of rare cultural entities. Similarly, I-HallA~\cite{lim2025evaluating} employs iterative VQA procedures together with external large language models (LLMs) to detect geographical, folkloric, or culturally specific hallucinations. These methods provide valuable tools for auditing cultural and factual errors in generated images, but they often require retrieval, multi-step reasoning, or repeated model queries.

\noindent\textbf{Fine-grained cultural feature extraction.}
A closely related work, \textit{Culture in Action}~\cite{cultivate2026}, introduces the CULTIVate benchmark and the AHEaD framework for evaluating T2I models through culturally grounded social activities. AHEaD is similar to our work in its focus on identifying localized, culturally meaningful details during evaluation. However, it relies on explainable text descriptors and iterative MLLM-based reasoning to assess cultural alignment, hallucination, and exaggeration. In contrast, our goal is to learn a lightweight scalar reward model that avoids autoregressive rationale generation and can be used for efficient preference assessment and downstream alignment.

\noindent\textbf{The evaluation-to-alignment gap.}
Despite their strengths, many culture-specific evaluation frameworks are primarily designed as offline auditing pipelines rather than lightweight reward models for alignment. Their reliance on external retrieval, iterative VQA, or human-in-the-loop verification can make them expensive to use within large-scale preference optimization loops. This creates an evaluation-to-alignment gap: culturally aware evaluation methods can identify failures, but they are often difficult to convert into efficient reward signals for improving T2I models. Our work addresses this gap by introducing an Implicit Cultural Probe and a SkipCA mechanism that enable a lightweight MLLM-based reward model to revisit early visual representations and produce an efficient scalar cultural alignment score without external retrieval or autoregressive rationale generation.

\section{Defining Cultural Boundaries and Scope}

To make cultural bias evaluation tractable and reproducible, it is necessary to operationalize the abstract concept of ``culture'' within a clearly defined evaluation scope. In this work, we follow the theoretical framing and categorization protocol of the \textit{CulturalFrames} benchmark~\cite{nayak2025culturalframes} to define the cultural dimensions considered in our study. Rather than treating culture as a fixed or homogeneous identity, we use this framework to examine whether T2I evaluators can recognize culturally salient visual details that are relevant to a given prompt and context.

\noindent\textbf{Country as an operational proxy.}
Culture is inherently complex, fluid, and internally diverse. For quantitative evaluation, we use ``country'' as an operational proxy for culturally grounded contexts, following prior benchmark construction practices and broader discussions on measuring culture in AI systems~\cite{nayak2025culturalframes,adilazuarda2024towards,hofstede2010cultures}. This choice does not imply that a country corresponds to a single unified culture. Instead, it provides a practical unit for organizing prompts, annotations, and visual expectations around nationally or regionally salient practices, institutions, artifacts, and symbolic references. We therefore interpret country-level labels as evaluation contexts rather than essentialized cultural identities.

\noindent\textbf{Country selection strategy.}
To construct a diverse evaluation benchmark, our selected countries span multiple geographic regions and cultural contexts, following the cultural-zone categorization used in the World Values Survey (WVS)~\cite{haerpfer2022world} and the \textit{CulturalFrames} benchmark. The selection includes India, China, Japan, Iran, South Africa, Brazil, Chile, Canada, Poland, and Germany, covering a broad range of Asian, African, Latin American, European, and English-speaking contexts. This design allows us to evaluate whether T2I reward models can generalize beyond dominant Western visual conventions and recognize culturally specific details across underrepresented and non-Western settings.

\noindent\textbf{Five sociocultural domains.}
To identify and evaluate implicit cultural expectations, our study considers five sociocultural domains derived from the \textit{CulturalAtlas} framework~\cite{mosaica2024culturalatlas}. These domains are commonly reflected in everyday visual scenes, social practices, and ceremonial contexts, making them suitable dimensions for diagnosing whether generated images preserve culturally meaningful details:
\begin{enumerate}
\item \textit{Family:} Familial roles, intergenerational relationships, household structures, and interaction norms.
\item \textit{Greetings:} Customary practices in social, formal, and business interactions.
\item \textit{Etiquette:} Everyday behavioral norms, including visits, meals, gift-giving, and public conduct.
\item \textit{Religion:} Religious rituals, ceremonial practices, sacred objects, and worship-related contexts.
\item \textit{Dates of Significance:} Celebrations, commemorations, festivals, and events of cultural, historical, or religious importance.
\end{enumerate}

We clarify that cultural alignment is defined as prompt-conditioned cultural plausibility rather than a fixed notion of cultural correctness. Following CulturalFrames and related cultural competence frameworks such as CUBE, we treat culturally appropriate representations as those containing visual artifacts, practices, or contextual details that are broadly plausible for the specified prompt and cultural context. This definition does not assume a single canonical representation of any culture; instead, it allows intra-cultural variation while distinguishing plausible alternatives from explicit misalignments, missing culturally salient details, or hallucinated artifacts.

\begin{figure}[t]
\centering
\includegraphics[width=0.9\textwidth]{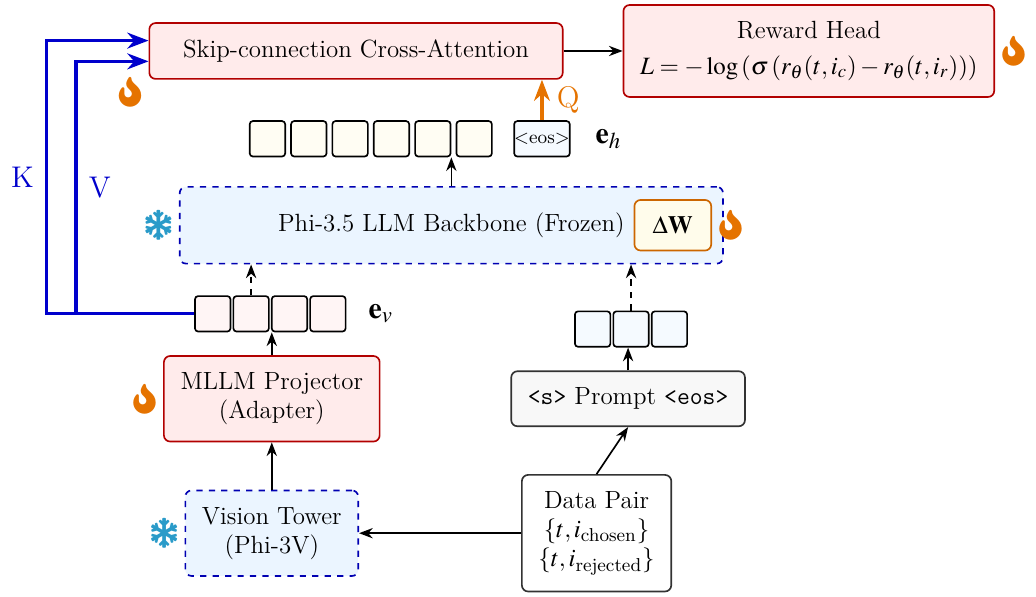}
\caption{Overview of the proposed Implicit Cultural Alignment Reward Model. The model uses the Phi-3.5-vision backbone to process text-image pairs together with an implicit cultural probe. A SkipCA module enables the final EOS hidden representation to revisit early visual tokens before producing a scalar reward. The model is optimized using a pairwise Bradley-Terry ranking loss.}
\label{fig:architecture}
\end{figure}

\subsection{Architecture Design}
As illustrated in Figure~\ref{fig:architecture}, we adopt the 4.2B-parameter Phi-3.5-vision~\cite{abdin2024phi3} as our base MLLM, following the Phi family of compact language models, due to its compact model size and strong vision-language representation capability. We adapt this backbone by introducing a cross-attentive reward head for scalar reward prediction.

Unlike VQA-based evaluators that require MLLMs to generate textual rationales or categorical judgments, our method bypasses autoregressive decoding during inference. Instead, it directly maps multimodal hidden representations to a continuous reward scalar. This design avoids lengthy human-crafted evaluation prompts and substantially improves inference efficiency, making the model more suitable for large-scale evaluation and preference optimization.

\subsection{Implicit Cultural Probe}
To capture implicit cultural expectations, we derive a lightweight probe from the \textit{CulturalAtlas} framework. Unlike VQA evaluators requiring task-specific prompts, our probe employs a single fixed guideline across all samples, introducing negligible overhead. It provides contextual guidance across five sociocultural dimensions (Family, Greetings, Etiquette, Religion, and Dates of Significance), encouraging the model to evaluate meaningful cultural cues rather than mere object presence.

Specifically, the probe acts as a fixed system instruction prepended to the user prompt before tokenization. It directs the model with a concise mandate:

\vspace{2mm}
\noindent\fbox{
    \begin{minipage}{\dimexpr\linewidth-2\fboxsep-2\fboxrule\relax}
    \small \texttt{[System Instruction]} \\
    Task: Cultural Authenticity Assessment. \\
    Evaluate how well the image aligns with the explicit text prompt and its expected implicit cultural context. Pay attention to contextually accurate details across relevant socio-cultural dimensions, including: family dynamics, social greetings, dining etiquette, religious practices, and festive settings. \\Assess the presence of authentic local artifacts and norms, penalizing culturally incongruous elements or generic visual stereotypes.
    \end{minipage}
}
\vspace{1mm}

This standardized prefix seamlessly steers the attention mechanism toward subtle cultural norms without requiring sample-specific prompt engineering or additional retrieval.

\subsection{SkipCA Reward Head}
In contemporary LLM-based reward modeling, the reward head is often implemented as a simple linear projection layer applied to the final hidden state. However, for implicit cultural alignment, relying solely on the final hidden representation may be insufficient because fine-grained visual evidence can be compressed or weakened in deeper transformer layers. Such details are particularly important when cultural alignment depends on small objects, ritual elements, utensils, clothing, gestures, or spatial arrangements.

To address this limitation, we replace the standard linear projection with a SkipCA reward head. Let $e_v \in \mathbb{R}^{n \times d}$ denote the projected visual tokens extracted immediately after the visual projector, where $n$ is the number of visual tokens and $d$ is the hidden dimension. Let $e_h \in \mathbb{R}^{1 \times d}$ denote the hidden state of the end-of-sequence token from the final MLLM layer~\cite{vaswani2017attention}. SkipCA uses $e_h$ as the query and $e_v$ as the key and value:
\begin{equation}
Q = e_h W_Q, \quad K = e_v W_K, \quad V = e_v W_V,
\end{equation}
\begin{equation}
f_{\mathrm{SkipCA}}(e_h,e_v) = \mathrm{softmax}\left(\frac{QK^{\top}}{\sqrt{d_k}}\right)V,
\end{equation}
where $W_Q, W_K, W_V \in \mathbb{R}^{d \times d_k}$ are learnable projection matrices and $d_k$ denotes the attention dimension. The final scalar reward is then computed by applying a linear projection layer $l(\cdot)$ to the SkipCA output:
\begin{equation}
r_\theta(t,i) = l\left(f_{\mathrm{SkipCA}}(e_h,e_v)\right).
\end{equation}
By extracting $e_v$ immediately after the visual projector, SkipCA preserves access to early visual representations that contain more localized visual evidence. Cross-attending these visual tokens with the final end-of-sequence (EOS) hidden representation $e_h$ enables the reward head to selectively revisit culturally relevant details before producing the final reward score.

\subsection{Model Adaptation and Preference Objective}
We adapt the MLLM-based reward model using parameter-efficient fine-tuning. Specifically, we freeze the visual encoder and the original language-model backbone, while training the visual projector, the SkipCA reward head, and Low-Rank Adaptation (LoRA)~\cite{hu2022lora} modules $\Delta W$ inserted into the language model. This configuration keeps the number of trainable parameters small while allowing the model to adapt to culturally relevant multimodal preference signals.

For supervision, we use pairwise T2I preference data $\{t,i_c,i_r\}$, where $t$ is the text prompt, $i_c$ is the culturally preferred image, and $i_r$ is the rejected image generated from the same prompt. The model is trained with the Bradley-Terry pairwise ranking objective:
\begin{equation}
\mathcal{L} = -\log \sigma\left(r_{\theta}(t,i_c)-r_{\theta}(t,i_r)\right),
\end{equation}
where $\sigma(\cdot)$ denotes the sigmoid function. This objective encourages the reward assigned to the culturally preferred image to be higher than that assigned to the rejected image. By directly predicting scalar rewards from hidden representations, the proposed model learns a fine-grained preference signal for implicit cultural alignment without the computational overhead of autoregressive decoding.

\section{Experiments}

\subsection{Dataset and Evaluation Metrics}
To evaluate the cultural alignment performance of the proposed reward model, we use curated pairwise samples derived from the \textit{CulturalFrames} benchmark. Each sample consists of a text prompt $t$ and two generated images: a culturally preferred image $i_c$ and a rejected image $i_r$ that is less aligned with the implicit cultural expectation associated with the prompt.

Following recent evaluations of multimodal reward models, we use \textit{pairwise accuracy} as the primary metric. Pairwise accuracy measures whether an evaluator assigns a higher reward score to the culturally preferred image than to the rejected image. Given a reward function $r_\theta(\cdot)$, a pair is considered correctly ranked if
\[
r_\theta(t,i_c) > r_\theta(t,i_r).
\]
In addition, to assess agreement with human cultural-alignment judgments, we report Pearson~\cite{pearson1895note} and Kendall~\cite{kendall1938new} correlation coefficients between model scores and human annotation scores.

\subsection{Dataset Construction and Data Splits}
Our experiments use augmented image pairs derived from the \textit{CulturalFrames} benchmark. The images were generated by four T2I models, including two open-source models (FLUX.1-dev~\cite{blackforest2024flux} and Stable Diffusion 3.5 Large~\cite{rombach2022high}) and two closed-source models (Imagen 3~\cite{google2024imagen3} and GPT-Image~\cite{betker2023improving}).

To ensure clear preference signals, we filter out pairs with small human-score margins (removing $\sim$9\%, retaining 3,323 pairs). Such small-margin pairs often reflect ambiguous cultural preferences or high visual similarity, which complicates pairwise reward learning. This filtering reduces label noise and focuses training on cases with strong human agreement. The dataset is partitioned into training, validation, and testing splits with an approximate ratio of 70\%, 15\%, and 15\%, respectively. The model is trained exclusively on the training split, the validation split is used for model selection, and the testing split is reserved solely for final evaluation.

The resulting pairs emphasize challenging cases involving implicit cultural expectations rather than solely prompt-explicit objects. This design encourages the model to learn culturally meaningful visual cues, including ritual elements, utensils, clothing, gestures, spatial arrangements, and other long-tail cultural artifacts.

\subsection{Implementation Details}
\label{subsec:impl_details}
To support reproducibility, we summarize our experimental configuration in Table~\ref{tab:implementation_details}. Our Implicit Cultural Alignment Reward Model is built upon \textit{Phi-3.5-vision-instruct}, which contains 4,419,635,200 parameters in our implementation. We apply parameter-efficient fine-tuning using LoRA: the core vision encoder and language-model weights are frozen, while the visual projector, the proposed SkipCA reward head, and the LoRA modules are trainable. This results in 307,638,272 trainable parameters, corresponding to approximately 6.96\% of the full network.

\begin{table}[t]
\centering
\caption{Detailed hyperparameter configurations for our reward model training.}
\label{tab:implementation_details}
\scriptsize
\begin{tabular}{lc}
\toprule
\textbf{Hyperparameter} & \textbf{Value} \\
\midrule
\multicolumn{2}{c}{\textit{Model Architecture}} \\
\midrule
Base Model Backbone & Phi-3.5-vision-instruct (4.2B) \\
LoRA Rank ($r$) & 128 \\
LoRA Alpha ($\alpha$) & 256 \\
LoRA Dropout & 0.1 \\
Target Modules & all-linear \\
Total Parameters & 4,419,635,200 \\
Trainable Parameters & 307,638,272 ($6.9607\%$) \\
\midrule
\multicolumn{2}{c}{\textit{Training Configurations}} \\
\midrule
Optimization & AdamW \\
Learning Rate & $1 \times 10^{-4}$ \\
Training Epochs & 3 \\
Micro-train Batch Size & 4 \\
Gradient Accumulation Steps & 4 \\
Global Batch Size & 16 \\
Max Sequence Length & 2048 \\
Distributed Framework & DeepSpeed ZeRO-2 \\
Hardware Acceleration & Flash Attention 2 \\
\midrule
\multicolumn{2}{c}{\textit{Hardware \& Evaluation Setup}} \\
\midrule
GPU Infrastructure & Single NVIDIA RTX 5090 (32GB) \\
Host CPU Specification & Intel Core Ultra 7 265K \\
Image Patch Resolution & $336 \times 336$ (Dynamic Multi-crop) \\
Inference Batch Size & 1 (Per forward pass) \\
\bottomrule
\end{tabular}
\end{table}

\subsection{Performance Analysis}
Table~\ref{tab:performance} reports the quantitative comparison between our model and representative T2I evaluators. Our model achieves the best performance among the compared methods, reaching 83.49\% pairwise accuracy. It outperforms VQAScore at 76.83\% and GPT-4o at 72.54\%. In terms of correlation with human cultural-alignment judgments, our model also achieves the highest Pearson correlation coefficient of 0.5268 and Kendall coefficient of 0.3749. These results indicate that the proposed reward model provides a more reliable scalar signal for implicit cultural alignment than general-purpose vision-language or MLLM-based evaluators.

\begin{table}[t]
\centering
\caption{Quantitative evaluation of cultural alignment performance. Best results are highlighted in bold.}
\label{tab:performance}
\scriptsize
\begin{tabular}{@{}lccc@{}}
\toprule
Method & Acc. (\%) & Pearson & Kendall \\
\midrule
CLIPScore & 52.60 & 0.080 & 0.060 \\
PickScore & 67.96 & 0.223 & 0.146 \\
GPT-4o & 72.54 & 0.405 & 0.267 \\
VQAScore & 76.83 & 0.432 & 0.363 \\
\textbf{Our Model} & \textbf{83.49} & \textbf{0.5268} & \textbf{0.3749} \\
\bottomrule
\end{tabular}
\end{table}

\subsection{Qualitative Analysis}
\label{subsec:qualitative}

\begin{figure}[t]
\centering
\renewcommand{\arraystretch}{1.2}

\begin{tabular}{@{} p{0.27\linewidth} p{0.27\linewidth} p{0.42\linewidth} @{}}
\toprule
\centering \textbf{Chosen} & \centering \textbf{Rejected} & \textbf{Rejection Rationale} \\
\midrule

% --- 第一組：Explicit Issue ---

\multicolumn{3}{@{}p{0.96\linewidth}@{}}{\textbf{Case 1 (Explicit Issue):} ``Priest blessing herbs in German countryside church''} \\
\vspace{0pt} \centering \includegraphics[width=1\linewidth,height=1\linewidth]{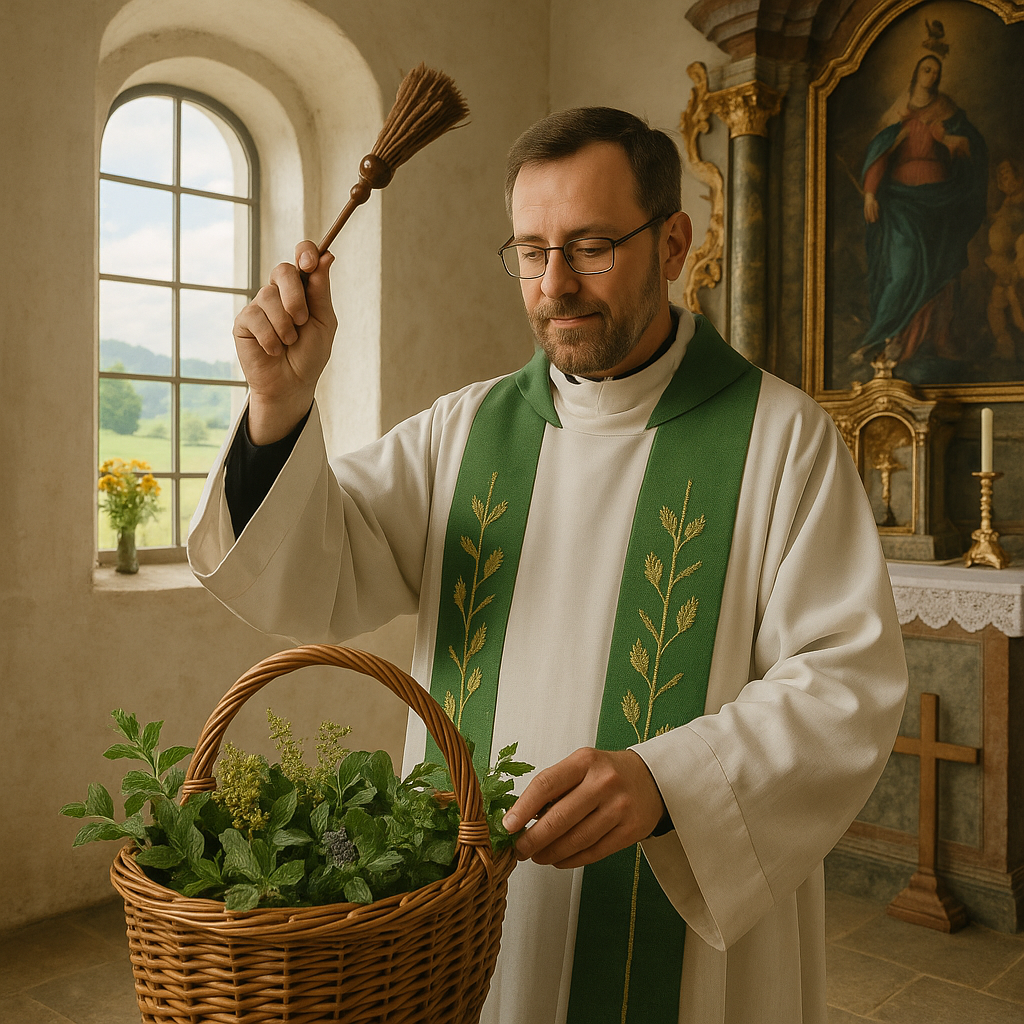} \par \vspace{1pt} \small{Reward: +1.22} &
\vspace{0pt} \centering \includegraphics[width=1\linewidth,height=1\linewidth]{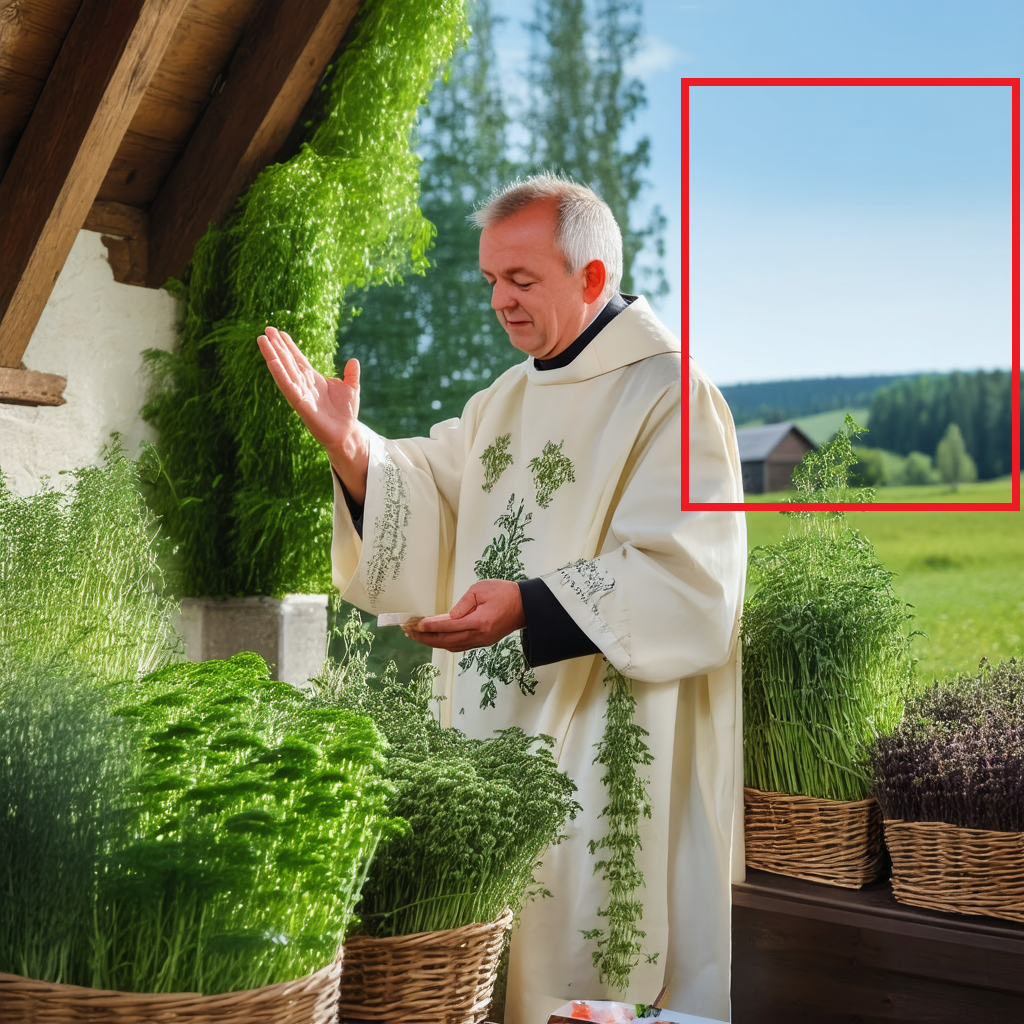} \par \vspace{1pt} \small{Reward: +0.51} &
\vspace{0pt} \small{``Does not look like a church or no church is shown.''} \\
\midrule

% --- 第二組：Implicit Issue ---
\multicolumn{3}{@{}p{0.96\linewidth}@{}}{\textbf{Case 2 (Implicit Issue):} ``Children playing traditional taiko drums at a Japanese summer festival''} \\
\vspace{0pt} \centering \includegraphics[width=1\linewidth,height=1\linewidth]{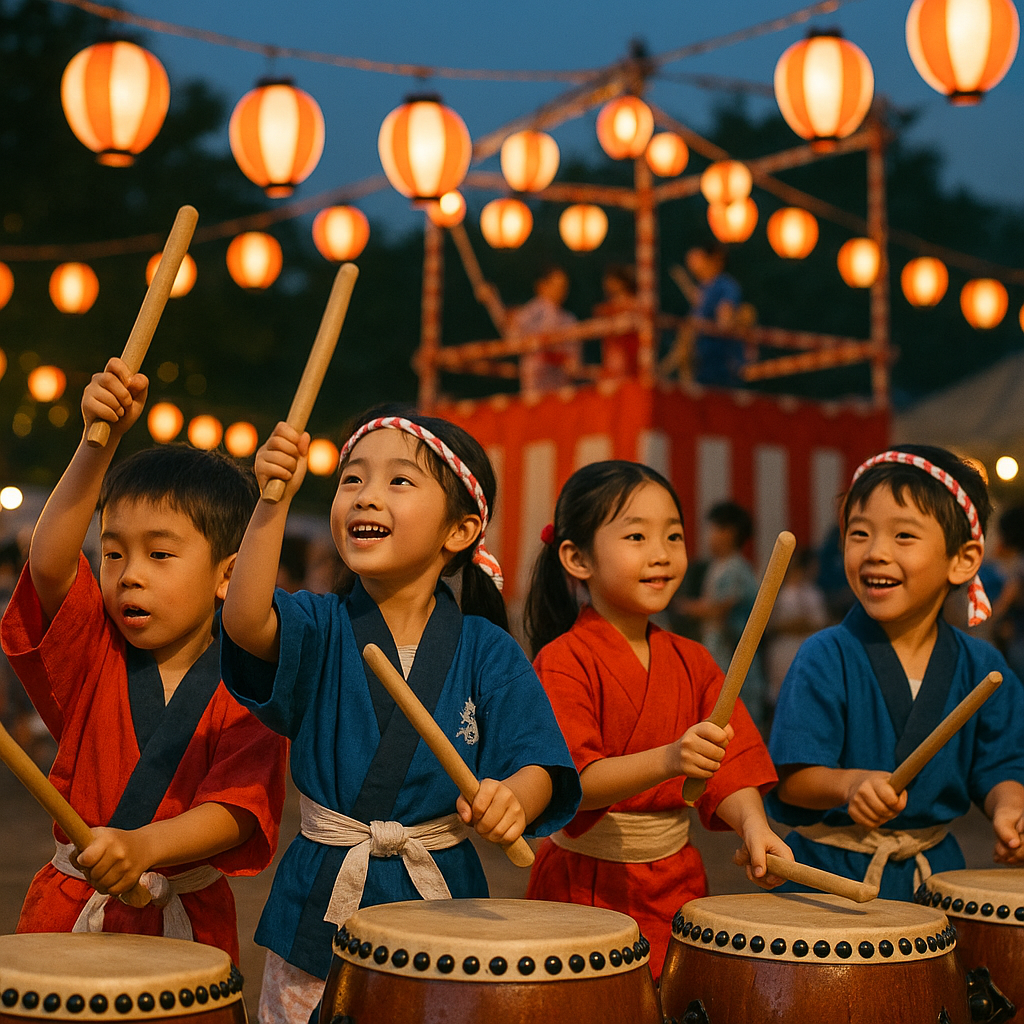} \par \vspace{1pt} \small{Reward: +1.14} &
\vspace{0pt} \centering \includegraphics[width=1\linewidth,height=1\linewidth]{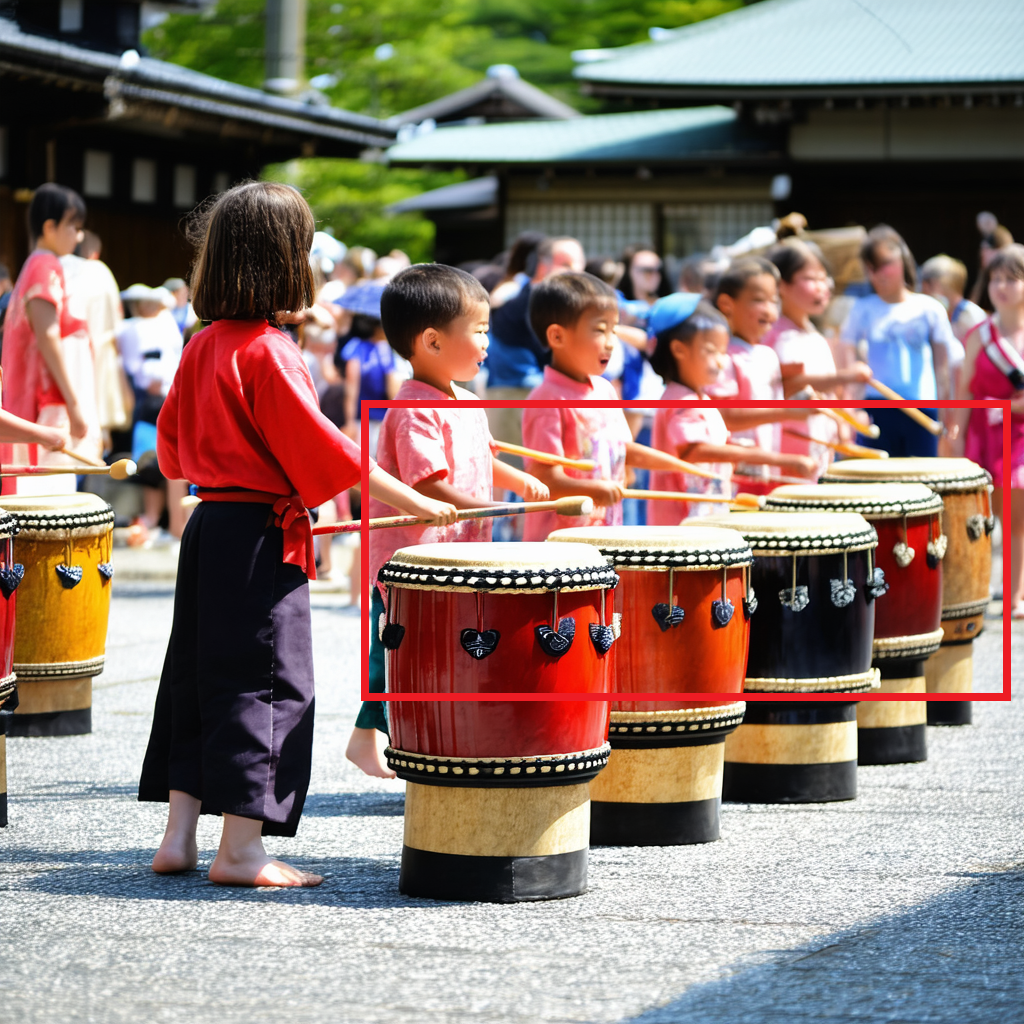} \par \vspace{1pt} \small{Reward: +0.44} &
\vspace{0pt} \small{``Unnatural taiko drum, the sticks don't hit taiko drums.''} \\
\midrule

% --- 第三組：Both Issue ---
\multicolumn{3}{@{}p{0.96\linewidth}@{}}{\textbf{Case 3 (Compound Issue):} ``Friends enjoying feijoada together at a Brazilian home.''} \\
\vspace{0pt} \centering \includegraphics[width=1\linewidth,height=1\linewidth]{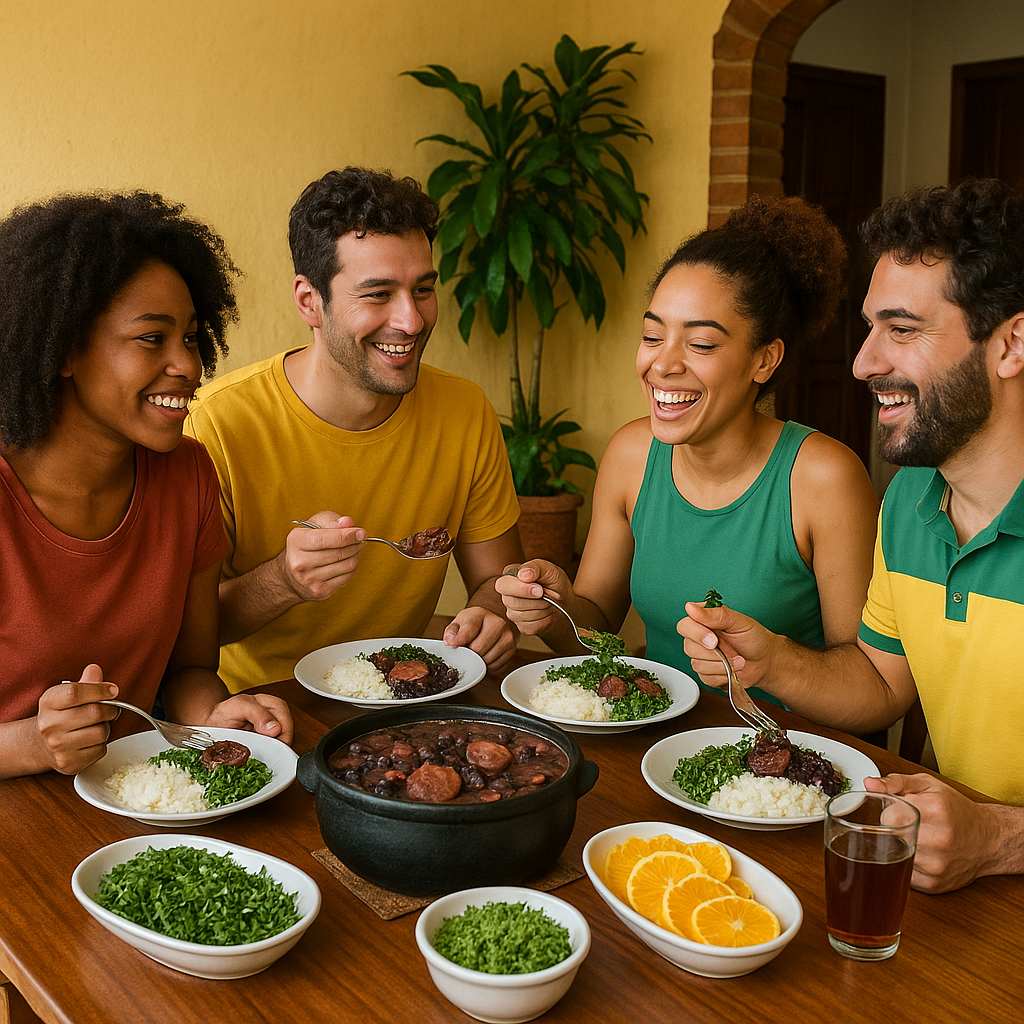} \par \vspace{1pt} \small{Reward: +1.38} &
\vspace{0pt} \centering \includegraphics[width=1\linewidth,height=1\linewidth]{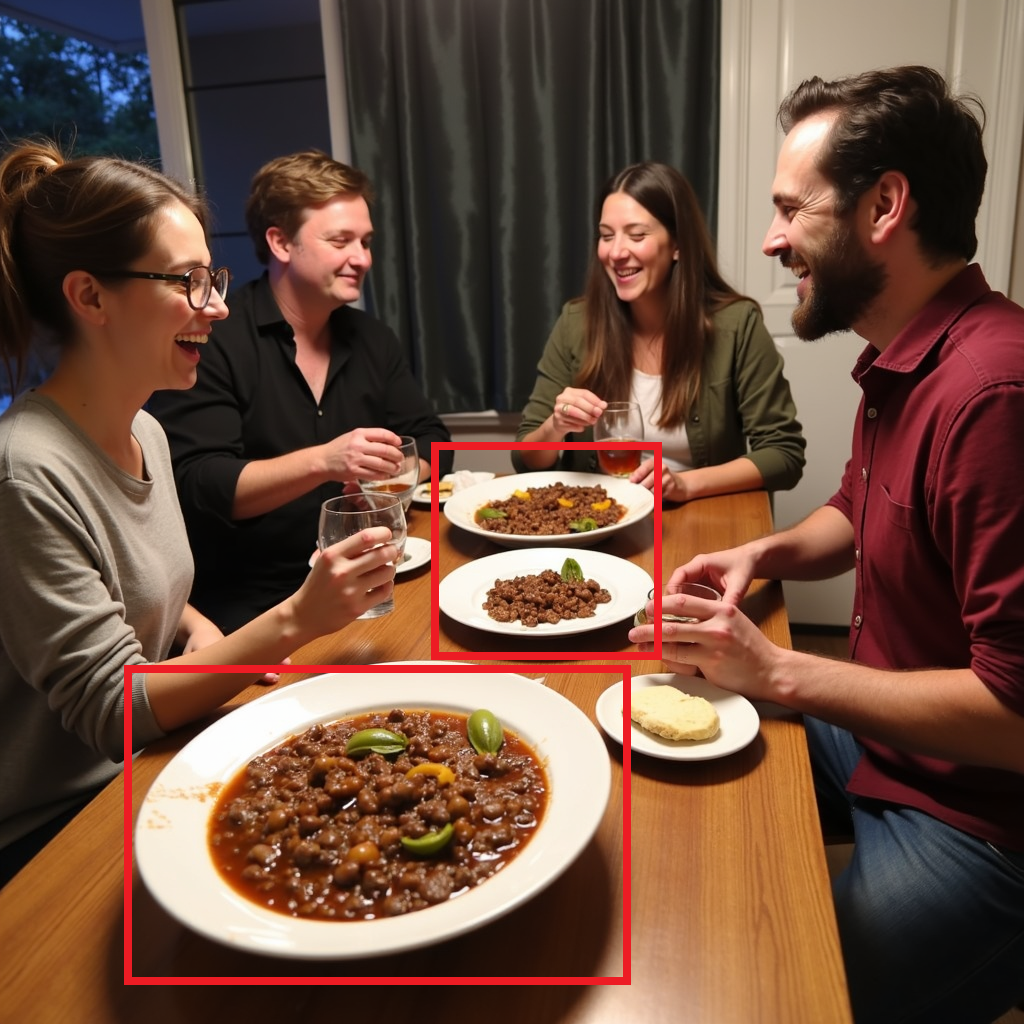} \par \vspace{1pt} \small{Reward: +0.87} &
\vspace{0pt} \small{``The environment is not typical Brazilian. The image shows a brownish liquid with green grains that resemble peas, which is inaccurate. A traditional Brazilian feijoada is made with black beans, and it does not include green peas.''} \\

\bottomrule
\end{tabular}
\caption{Qualitative pairs demonstrating three distinct baseline failure modes. Our model's continuous scalar rewards are mapped directly underneath each sub-image, closely aligning with the nuanced critiques provided by native human annotators.}
\label{fig:qualitative_pairs}
\end{figure}

To further examine how the proposed reward model aligns with human cultural-alignment judgments, we present three prompt-conditioned image pairs in Figure~\ref{fig:qualitative_pairs}. Each example compares a culturally preferred image with a rejected image, together with the scalar rewards predicted by our model and the corresponding human annotation rationale. The examples cover three types of failures: explicit prompt mismatch, implicit cultural mismatch, and combined explicit-implicit mismatch.

As shown in Figure~\ref{fig:qualitative_pairs}, our model assigns lower scores to images flagged by human annotators. In Case 1, the rejected image fails to depict the explicit target setting, resulting in a lower reward. In Case 2, the rejected image contains a subtle implicit issue involving an unnatural depiction of traditional Japanese taiko drums. In Case 3, the rejected image contains both environmental misalignment and a food-related cultural inconsistency. These qualitative examples suggest that the learned reward function is sensitive to localized visual-semantic cues that are relevant to prompt-conditioned cultural authenticity.

\subsection{Inference Efficiency}
In addition to cultural-alignment performance, inference efficiency is essential for using reward models in large-scale evaluation or preference optimization. As shown in Table~\ref{tab:efficiency}, our model processes each image in 0.21 seconds by directly predicting a scalar reward without relying on autoregressive text generation. Compared with the VQA-based VQAScore, which requires 2.82 seconds per image, our model achieves approximately a $10\times$ speedup while using a smaller 4.2B-parameter backbone.

Although CLIPScore and PickScore are faster due to their compact embedding-based architectures, their cultural-alignment performance is substantially lower, as shown in Table~\ref{tab:performance}. This suggests that our model provides a favorable trade-off between cultural-alignment accuracy and inference efficiency. For GPT-4o, we report API latency rather than local inference time; therefore, its timing is not directly comparable with locally deployed models.

\begin{table}[t]
\centering
\caption{Efficiency comparison. GPT-4o reports API latency rather than local inference time.}
\label{tab:efficiency}
\renewcommand{\arraystretch}{1.2}
\scriptsize
\begin{tabular}{@{}lcc@{}}
\toprule
Method & Parameters & Inference Time (s) \\
\midrule
CLIPScore & 428M & 0.01 \\
PickScore & 986M & 0.06 \\
GPT-4o & -- & 0.42\textsuperscript{*} \\
VQAScore & 11B & 2.82 \\
\midrule
\textbf{Ours} & \textbf{4.42B} & \textbf{0.21} \\
\bottomrule
\end{tabular}

\vspace{2pt}
\footnotesize{\textsuperscript{*}GPT-4o latency is measured through API access and is not directly comparable to local inference time.}
\end{table}

\subsection{Ablation Study}
To systematically evaluate the contribution of each proposed component, we conduct an ablation study on the testing split of the augmented \textit{CulturalFrames} dataset. We compare the Full Model against three variants:
\begin{itemize}
    \item \textbf{Variant A (w/o Cultural Prompt):} The implicit cultural probe is removed during inference, and the model relies only on the explicit user prompt.
    \item \textbf{Variant B (w/o SkipCA):} The SkipCA mechanism is removed, and the reward is predicted directly from the final EOS hidden state using a standard linear projection.
    \item \textbf{Variant C (Frozen Projector):} The visual projector is frozen during training, limiting the adaptation of early visual features to the cultural reward modeling objective.
\end{itemize}

The quantitative results are summarized in Table~\ref{tab:ablation}. The Full Model achieves the best performance overall, demonstrating that the implicit cultural probe, SkipCA reward head, and trainable visual projector jointly contribute to cultural alignment evaluation. Compared with the Full Model, Variant B (w/o SkipCA) and Variant C (Frozen Projector) both show lower accuracy and correlation, with Variant C exhibiting a larger accuracy drop, while the two variants achieve closely comparable Pearson and Kendall scores, suggesting the visual projector contributes more to ranking decisions while SkipCA contributes more evenly across metrics. Variant A (w/o Cultural Prompt) performs worst overall, highlighting the probe's role in guiding culturally salient evidence. These results indicate that our final architecture provides a more balanced reward signal for both pairwise preference prediction and continuous cultural-alignment scoring.

\begin{table}[t]
\centering
\caption{Ablation study on the testing split of the augmented CulturalFrames dataset. Best results are highlighted in \textbf{bold}, and the second-best results are \underline{underlined}.}
\label{tab:ablation}
\scriptsize
\begin{tabular}{@{}lccc@{}}
\toprule
Model & Acc. (\%) & Pearson & Kendall \\
\midrule
Variant A (w/o Cultural Prompt) & 74.68 & 0.4286 & 0.3173 \\
Variant B (w/o SkipCA) & \underline{81.10} & \underline{0.4686} & 0.3365 \\
Variant C (Frozen Projector) & 77.43 & 0.4657 & \underline{0.3368} \\
\textbf{Full Model} & \textbf{83.49} & \textbf{0.5268} & \textbf{0.3749} \\
\bottomrule
\end{tabular}
\end{table}

\section{Limitations}
While our proposed model demonstrates strong performance and efficient inference, several limitations remain. First, our current evaluation is bounded by the scope of the \textit{CulturalFrames} benchmark. Although country-level labels provide a tractable proxy for organizing cultural contexts, they are necessarily coarse and cannot fully capture the diversity, hybridity, and intersectionality of cultural identities and practices. Future work should incorporate a broader spectrum of underrepresented communities and adopt more dynamic, fine-grained cultural characterizations that extend beyond rigid geopolitical boundaries.

Because our evaluation uses image pairs filtered for clear preference margins to reduce noise, generalization to subtle or low-agreement cross-cultural cases remains limited. Furthermore, the baseline evaluators were not explicitly optimized for this filtered distribution, which may partially account for the performance gap. Thus, our results demonstrate effectiveness within a strong-signal benchmark rather than a guarantee of robustness across all ambiguous real-world scenarios.

Second, our model is designed as an evaluation and reward modeling framework rather than a direct generator debiasing method. Although the learned reward signal may support downstream preference optimization, this paper does not fine-tune T2I generators or evaluate whether the proposed reward model improves generation quality after RLHF or DPO. Establishing this full evaluation-to-alignment pipeline remains an important direction for future work.

Third, to balance inference efficiency and deployment feasibility, we adopt a lightweight 4.2B-parameter backbone. Although the SkipCA mechanism enables the reward head to revisit early visual representations, the model's capacity to interpret exceptionally rare or long-tail cultural artifacts remains constrained by the internal visual and cultural knowledge of the base MLLM. Incorporating external knowledge sources or retrieval-augmented cultural grounding may further improve robustness in such cases.

\section{Conclusion and Future Work}
In this paper, we introduced an Implicit Cultural Alignment Reward Model for evaluating T2I generation under implicit cultural expectations. By integrating an implicit cultural probe with a SkipCA reward head, the proposed framework enables the final semantic representation of a lightweight MLLM to revisit early-stage visual tokens before producing a scalar cultural alignment score. This design improves sensitivity to fine-grained cultural artifacts while avoiding the inference cost of autoregressive VQA-style evaluation.

Experiments on 3,323 curated pairwise samples derived from the \textit{CulturalFrames} benchmark show that our model achieves 83.49\% pairwise accuracy, outperforming representative evaluators such as GPT-4o and VQAScore. The model also achieves stronger agreement with human cultural-alignment judgments, with Pearson and Kendall coefficients of 0.5268 and 0.3749, respectively. In addition, by directly predicting scalar rewards without autoregressive decoding, our method reduces average inference latency to 0.21 seconds per image, providing an efficient reward signal for large-scale multimodal preference assessment.

Future work will extend this framework in two directions. First, we will broaden the cultural scope of the benchmark to include more fine-grained and underrepresented cultural contexts. Second, we will investigate the downstream use of the proposed reward model for preference optimization of T2I generators, including RLHF- and DPO-style training pipelines. These extensions will help determine whether culturally aware reward modeling can not only evaluate generated images but also improve the cultural reliability and trustworthiness of generative systems in practice.

\paragraph{Acknowledgements.}
This work was supported by the MOE Yushan Young Scholar Program under Grant MOE-114-YSFEE-0010-008-P1 and NSTC Taiwan under Grant NSTC 115-2813-C-A49-146-E.

\clearpage

\bibliographystyle{splncs04}
\bibliography{main}
\end{document}